\documentclass[10pt, a4paper]{article}

\usepackage{xcolor} 
\usepackage{lrec-coling2024} 

\usepackage{enumitem}
\usepackage{url}
\usepackage{amssymb}
\usepackage{adjustbox}
\usepackage{multirow}
\usepackage{booktabs}
\usepackage[justification=centering]{subfig}
\graphicspath{ {./img/} }
\usepackage{array}
\newcolumntype{H}{>{\setbox0=\hbox\bgroup}c<{\egroup}@{}}

\title{Prompting-based Synthetic Data Generation\\for Few-Shot Question Answering}

\name{Maximilian Schmidt\textsuperscript{1}, Andrea Bartezzaghi\textsuperscript{2}, Ngoc Thang Vu\textsuperscript{1}} 

\address{\textsuperscript{1} University of Stuttgart, \textsuperscript{2}IBM Research Zurich \\
\textsuperscript{1}\{maximilian.schmidt, thang.vu\}@ims.uni-stuttgart.de, \textsuperscript{2}abt@zurich.ibm.com\\}

\abstract{
  Although language models (LMs) have boosted the performance of Question Answering, they still need plenty of data.
  Data annotation, in contrast, is a time-consuming process.
  This especially applies to Question Answering, where possibly large documents have to be parsed and annotated with questions and their corresponding answers.
  Furthermore, Question Answering models often only work well for the domain they were trained on.
  Since annotation is costly, we argue that domain-agnostic knowledge from LMs, such as linguistic understanding, is sufficient to create a well-curated dataset.
  With this motivation, we show that using large language models can improve Question Answering performance on various datasets in the few-shot setting compared to state-of-the-art approaches.
  For this, we perform data generation leveraging the Prompting framework, suggesting that language models contain valuable task-agnostic knowledge that can be used beyond the common pre-training/fine-tuning scheme.
  As a result, we consistently outperform previous approaches on few-shot Question Answering.
 }

\begin{document}

\maketitleabstract

\section{Introduction}

Machine Reading Question Answering (MRQA) is an important task in Natural Language Processing and allows to easily access information by providing answers to specific questions.
While there are several subtasks related to MRQA such as open-domain, binary/multiple choice, conversational or generative QA, we focus on extractive QA  in this work.
In extractive QA, the goal is to find the answer to a question by extracting it from a given context.
MRQA has also raised attention in the community as a surrogate where other tasks are cast as question answering problems, thereby enabling a broad range of applications.
This includes, for example, Named Entity Recognition (NER, \citealp{li_unified_2020,arora_split-ner_2023}), entity relation extraction \citep{levy_zero-shot_2017,li_entity-relation_2019,zhang_entqa_2022} and slot filling \citep{gao_dialog_2019}.

Pre-training language models (LMs) on Natural Language Understanding (NLU) objectives such as Masked Language Modeling (MLM, \citealp{devlin_bert_2019}) led to strong MRQA models \citep{rajpurkar_squad_2016} and even surpasses human level\footnote{e.g., on the SQuAD \citep{rajpurkar_squad_2016} benchmark: \url{https://rajpurkar.github.io/SQuAD-explorer/}}.
Since the downstream task uses a different objective function, fine-tuning pre-trained LMs (PLMs) is necessary to adapt to the task.
Arguably, this misalignment leads to poor results if labeled data for the downstream task is scarce.
However, annotating data for MRQA is time-consuming and expensive.
Additionally, few-shot MRQA poses an interesting challenge, especially for specific domains, where high effort is needed to annotate data or domain experts are missing.
Also, there is still a gap in the performance of few-shot models when compared to the high-resource setting.
For example, the best model on TextbookQA \citep{kembhavi_are_2017} using 16 labeled samples is currently reported in literature to reach at most 49.9\% F1 \citep{castel_how_2022}.

\begin{figure}[t]
    \centering
    \includegraphics[trim=10 0 0 0,clip,width=.9\linewidth,keepaspectratio]{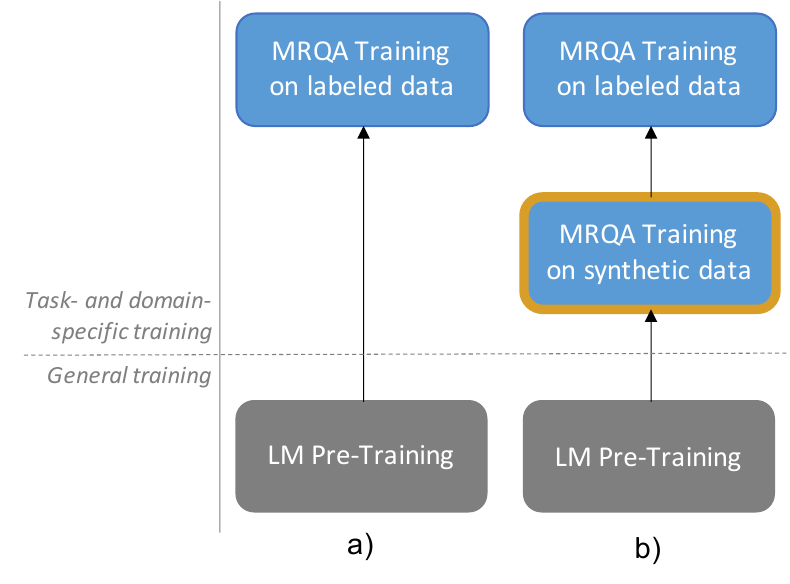}
    \caption{
        Comparison of a) common approaches, e.g., Prompting, for MRQA and b) our approach adding synthetic task- and domain-specific data without the need of additional labeled data.
    }
    \label{fig:idea}
\end{figure}

To deal with the low-resource MRQA setting, previous work has proposed to generate synthetic data to augment the training set \citep[e.g.,][]{alberti_synthetic_2019,puri_training_2020,shakeri_end--end_2020,shakeri_towards_2021}.
With a similar objective, PLMs have been employed for other tasks in Natural Language Processing \citep[e.g.,][]{anaby-tavor_not_2020,schick_exploiting_2021}.
However, MRQA is more challenging when generating synthetic data:
We cannot simply generate text given a label, but have to come up with a sample's input as well as its label in the form of a question and its answer.
Additionally, both the answer and the question are mutually dependent.
In this work, we explore an approach that enables this; a high-level overview is given in figure \ref{fig:idea}.

More precisely, we aim to answer the following research questions:
\begin{enumerate}[align=parleft, wide]
    \item[\textbf{RQ1:}] How can we use LMs to generate synthetic data for improving the few-shot MRQA task?
          \begin{enumerate}[align=parleft, wide, labelindent=20pt]
              \item[\textbf{RQ1.1:}] To what extend can synthetic data improve performance?
              \item[\textbf{RQ1.2:}] How does the answer selection affect the performance?
              \item[\textbf{RQ1.3:}] Do we need labeled data at all or can LMs generate helpful data out of the box?
          \end{enumerate}
    \item[\textbf{RQ2:}] How does the proposed approach generalize to other domains?
\end{enumerate}

To achieve this, we believe that there is a more effective way to employ LMs:
We propose to use the linguistic knowledge encoded in these models to generate synthetic data for the target domain to counteract the effect of data scarcity.
For this, we use the LM's ability to generate questions conditioned on the input, and we argue that this can easily be carried out in any target domain since we build on unsupervised PLMs.

In summary, our contributions are as follows:
1) We propose an approach generating valuable labeled data for the target domain by using the linguistic knowledge encoded in LMs\footnote{Our source code is available publicly here: \url{https://github.com/mxschmdt/mrqa-prompting-gen}};
2) we improve the performance of few-shot QA for many dataset sizes across various domains to further bridge the performance gap between the few-shot and the full data setting; 
3) we demonstrate the high quality of questions generated by our approach in a user study.

While introducing a new, strong approach outperforming many state-of-the-art approaches in few-shot MRQA, our model even outperforms the full data setting of TextbookQA with 64\% F1 with only 64 labeled samples.

\section{Related Work}
In this section, we review existing work related to our setting, i.e., few-shot, as well as applications for Prompting.

\subsection*{Low-Resource MRQA}
Although there is no agreement among research on how much data the few-shot setting may comprise \citep{hedderich_survey_2021}, the objective when dealing with few-shot settings is to reduce the cost- and time-expensive annotation effort which, depending on the domain, may require domain expert knowledge.
Furthermore, for some domains, it is a challenge by itself to find experts or other resources~\citep{otegi_conversational_2020}.

Several approaches deal with settings where the amount of data is constrained. 
Many of them adopt the unsupervised pre-training technique.
While \citet{ram_few-shot_2021} come up with a QA-specific pre-training objective, many others adapt the LM to the target domain using its pre-training objective \citep{zhang_multi-stage_2020,nishida_unsupervised_2020,pergola_boosting_2021,chen_gotta_2023}.

Although self-supervised LM pre-training is also considered data augmentation, there also exist several approaches that deal with task- and domain-specific data augmentation.
For example, training instances can be manipulated by performing operations on the input keeping the labels the same \citep{zhang_multi-stage_2020}, or new labeled data can by synthesized \citep{alberti_synthetic_2019,shakeri_end--end_2020,shakeri_towards_2021}.

When humans are actively engaged in model development, Active Learning becomes possible \citep{settles_active_2012,schmidt_improving_2022}.

\subsection*{(L)LMs, Prompting}
As mentioned in the beginning, Prompting \citep{liu_pre-train_2021} aims to improve downstream tasks by aligning the pre-training objective with the downstream objective.
There are also several works that employ Prompting for the few-shot setting \citep{liu_pre-train_2021,schick_exploiting_2021}.
Empirically, Prompting alleviates the need for labeled data \citep{radford_language_2019,brown_language_2020} and also boosts QA performance in the few-shot setting \citep{chada_fewshotqa_2021,castel_how_2022}.
For example, \citet{chada_fewshotqa_2021} and \citet{wang_kecp_2022} align the MRQA task with the pre-training objective by casting the context-question-answer tuples as answer reconstruction, where the answer is decoded using an LM from the context and the question.
\citet{castel_how_2022} adapt this method for extractive MRQA by only decoding from the given context, i.e., computing probabilities on all possible spans from the context.

Several approaches aim to improve Prompting:
For example, soft tokens \citep{liu_low-resource_2022,li_prefix-tuning_2021,zhong_factual_2021} can allow the model to better adapt to downstream tasks and input data, and demonstration learning \citep{gao_making_2021} can help the model perform well in few-shot settings.

Prompting can also further be used for data augmentation \citep{anaby-tavor_not_2020}.

\begin{figure*}[tb]
    \centering
    \includegraphics[width=\textwidth]{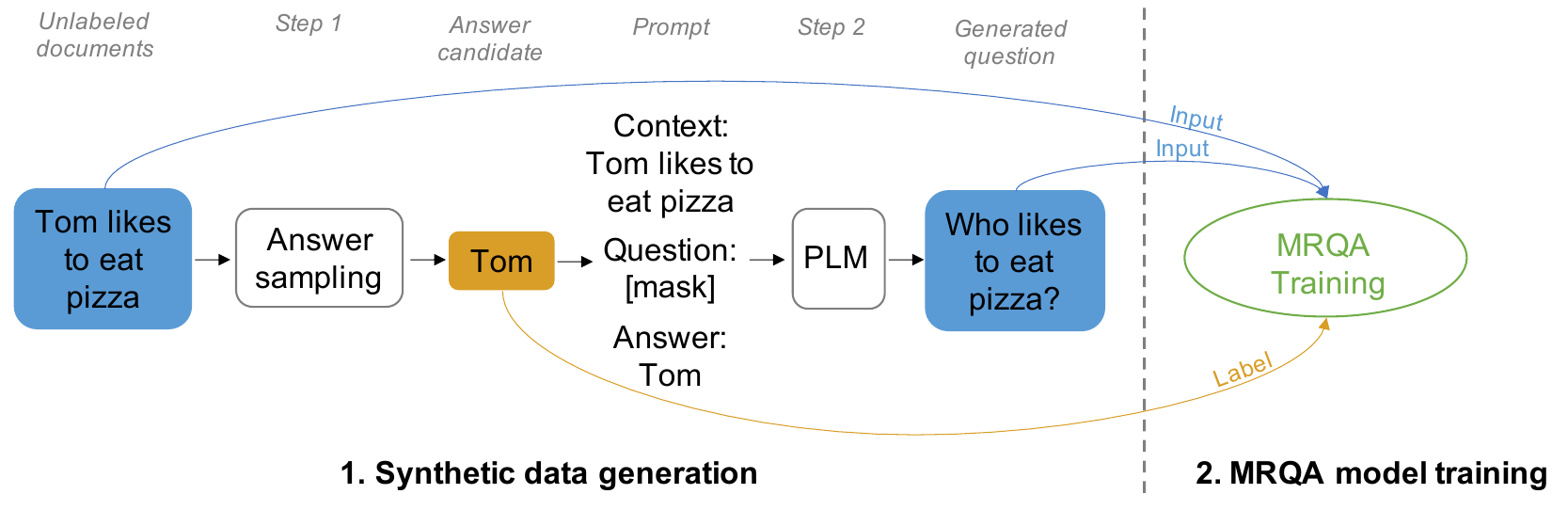}
    \caption{
        An example of our data generation pipeline:
        We first sample answer candidates (using NER) and then prompt a PLM to generate a question conditioned on context and answer (1).
        The generated question-answer pair is then used with the initial context to train an MRQA model (2).
        We afterwards perform additional training on labeled data if available.
    }
    \label{fig:method_example}
\end{figure*}

\section{Method}

Here, we give an overview of the problem and describe our approach in detail.

Formally, MRQA is defined as given context $c$ and question $q$, the goal is to predict the answer $a = f(c,q)$.
We further focus on extractive MRQA, that is, $a$ is a single contiguous span within $c$.

Next, we introduce our approach leveraging the Prompting framework.
The high-level idea of our approach is composed of two steps:
First, we sample answer candidates from a document.
In a second step, we then query a pre-trained LM for generating questions using the document and the previously sampled answers.
An overview of our data generation pipeline is given in figure \ref{fig:method_example}.
In the following we describe each step of o ur method in detail.

\subsection{Answer Sampling}
For sampling answer candidates, we apply NER\footnote{In fact, we also select values similar in style to names, e.g., periods.} to the context $c$ and the resulting entities are used as textual answers $a_c$ with spans $s$ (a tuple of character start and end indices).
We chose this technique because it is a simple resource-sparse approach and does not need to have knowledge on the domain's topic (i.e., any English NER model is sufficient for our domains).
Furthermore, NER is feasible in many languages and the datasets on which we evaluate our method, the few-shot MRQA benchmark, work with such style.
As a result, it can be applied to any domain for which a NER model exists in the given language.
Note that NER does not necessarily rely on many labeled samples or labeled data at all (e.g., rule-based approaches, using weak supervision \citep{lison_named_2020} or Prompting \citep{liu_low-resource_2022,ma_template-free_2022}).

\subsection{Question Generation}
\label{ssec:method_qg}
In Prompting, a LM takes a text input and, depending on the training objective, predicts the next token (as in the case of language modeling) or one or multiple masked tokens.
For example, T5 \citep{raffel_exploring_2020} is an encoder-decoder model which is fed a text input possibly containing multiple masked tokens.
For each masked token, one or more tokens can appear in the output prefixed by a \textit{sentinel} token, marking the masked token in the input to which the following tokens belong.
For our purpose we only use a single mask in the input.

For generating questions, we transform the sample inputs into prompts for the LM.
For this purpose we apply a template, thus replacing placeholders (marked starting with \textless{ }and ending with \textgreater) with the actual values from the sample.
Since we aim at generating a question given a context and a sampled answer candidate, the template is formulated to include the context and the answer, and the expected output is the question.
Therefore the question is formally defined as
\begin{equation}
    p(q|c,a_c) = \sum\limits_{t=1}^T \log p(q_t|q_{<t},c,a_c) .
\end{equation}

At training time, we use the original objective used for pre-training the underlying language model in order to model the question's probability by only computing the loss on the question $q$ in the output.
In our preliminary experiments, we have found that it is crucial to rely on sequence-to-sequence models, as these allow to condition the output not only on previous tokens but on the whole sequence.
We believe this is due to the more natural formulation where a question occurs before its answer in a sentence.
In contrast, using a left-to-right decoding model only, the input would have to be formulated such that the generated question can be answered by the previously (i.e., left to the question) given answer.
Obviously, this not only yields a longer prompt but also increases its complexity.
Furthermore, we make use of soft tokens in the input.
That is, all textual tokens from the template are trained in addition to the remaining model weights and we initialize them with the corresponding weights from the pre-trained word embedding.

For generation, we decode the question $q$ token-by-token and filtering is applied.
We do this for the following reasons:
1) the generated question could be noisy, e.g., not a valid question, and
2) the generated question may not be helpful for the Question Answering downstream task, possibly being underspecified.
For example, the generated question may have -- in addition to the provided answer -- several other correct answers in a given context.
As filtering technique we apply a two-step process.
First, we discard generated samples based on rule-based filtering.
We then apply consistency filtering \citep{alberti_synthetic_2019,anaby-tavor_not_2020} for which a model similar to the final MRQA model is employed.
Generated samples are discarded depending on the F1 score of the predicted answer (where the reference is the generated answer) using the MRQA model.
We do not use iterative consistency filtering \citep{wang_promda_2022} as this involves MRQA model re-training in each iteration, resulting in increased hyperparameter tuning complexity and resource drain.

\section{Experimental Setup}
Here, we first introduce the few-shot SQuAD dataset followed by implementational details of our approach and the description of baselines we consider.

\subsection{Few-shot Setting}

We perform experiments on several datasets in order to compare with existing approaches.
For this, we rely on the subsampled train and test splits from the Few-Shot MRQA benchmark\footnote{\url{https://github.com/oriram/splinter\#downloading-few-shot-mrqa-splits}} from \citet{ram_few-shot_2021} which are based on the preprocessed versions from the MRQA Shared Task 2019\footnote{\label{url:mrqa2019}\url{https://github.com/mrqa/MRQA-Shared-Task-2019}} \citep{fisch_mrqa_2019}.
More specific, this includes as domains SQuAD \citep{rajpurkar_squad_2016}, TriviaQA \citep{joshi_triviaqa_2017}, NaturalQuestionsShort (NQ) \citep{kwiatkowski_natural_2019}, NewsQA \citep{trischler_newsqa_2017}, HotpotQA \citep{yang_hotpotqa_2018}, BioASQ \citep{tsatsaronis_overview_2015} and TextbookQA \citep{kembhavi_are_2017} and we evaluate our approach using the splits with 16, 32, 64 and 128 training samples.
We perform our experiments for RQ1 on SQuAD while the remaining datasets are used to test the generalization capability of our approach (RQ2).

\subsection{Data Generation}
For generating questions, we have tested various models, templates and pre-processing strategies in preliminary experiments.
We found T5 \citep{raffel_exploring_2020} to perform best.
Instead of the original v1 model, which makes use of labeled data during pre-training thus violating our few-shot setting, we employ the v1.1 model in its large variant ($\sim$800M parameters).
Decoder-only models like GPT-2 \citep{radford_language_2019} performed worse as mentioned in section \ref{ssec:method_qg}.
In the templates, we considered case sensitivity as well as different wordings.
As a result of manual investigation, \textit{context: <context> question: <mask> answer: <answer>.}\footnote{\textless mask\textgreater is replaced by the model-specific mask token.} turned out to work well for our purpose and is similar to the findings of \citet{castel_how_2022}.

\subsubsection{Training of Question Generation Model}
For training the data generation model, similar to \citet{castel_how_2022} we create an academic development dataset to cater to the few-shot setting where having a separate development split leads to bad generalizability due to its small size.
Therefore, we tune the learning rate and the number of training steps on a validation set of 2048 samples from SQuAD's training data, and choose the set of hyperparameters that has the best normalized performance across all few-shot sizes as described in \citet{castel_how_2022}.
As a result, the question generation model was trained for 130 training steps with a batch size of 32 using a linear learning rate of 1e-4 with the Adafactor optimizer.
Furthermore, the soft tokens add 8192 weights for the question generation model.
We additionally chunk the provided contexts using a stride of 100 tokens so that at most 450 tokens of the context are included in a single input to allow for sufficient space for the question.
Since chunking can create instances where the answer is not part of the context, we drop these as we cannot expect them to yield a semantically correct question.

\subsubsection{Synthetic Data Generation}
For generating synthetic data, in case of SQuAD, TriviaQA, NQ, NewsQA, SearchQA and HotpotQA, we use the documents from the training corpus.
Since BioASQ and TextbookQA both comprise rather few documents, we collect abstracts from PubMed\footnote{\url{https://pubmed.ncbi.nlm.nih.gov/}} and lessons from CK-12\footnote{\url{https://www.ck12.org}}, respectively, for the purpose of generating data.
We then apply stanza's NER\footnote{\url{https://stanfordnlp.github.io/stanza/ner.html}} using all its entity types\footnote{\url{https://catalog.ldc.upenn.edu/docs/LDC2013T19/OntoNotes-Release-5.0.pdf} p.21f} in order to sample answers from these documents.
Afterwards, similar to training time, we apply chunking with a stride of 100 tokens to feed the documents into the model (by realizing the template) whereby we again only keep instances where the answer is contained in the context for the same reason as above.
In a subsequent step, the generated question is greedily decoded with a beam size of 5, top-k sampling with k equal to 20 and nucleus sampling \citep{holtzman_curious_2020} keeping tokens comprising 95\% of probability mass in each step.

The language models' special tokens are stripped in a subsequent step.
Since we only allow one mask in the input (for the question), we make sure that only the output tokens corresponding to this mask are used.

For the rule-based filtering in a subsequent step, we randomly select 1,000,000 samples and discard generated samples where the answer is contained in the question, or the question is only containing meaningless words or is empty.
Afterwards, we apply consistency filtering discarding generated samples with an F1 score of less than 80\% using a Prompting-based MRQA model which is trained similar to the one described in the next subsection.

\subsection{MRQA Model}
For the final step of our approach, we train an MRQA model using synthetic and available labeled data as shown in figure \ref{fig:method_example}.
Since a Prompting-based approach turned out to perform better than a span extraction head on top of a Transformer-based \citep{vaswani_attention_2017} encoder model, we also use T5 v1.1 (large) with the Prompting framework as our MRQA model.
Additionally, we compared T5 v1.1 pre-trained with and without recurring span selection (RSS)\footnote{\url{https://huggingface.co/tau/t5-v1_1-large-rss}} \citep{castel_how_2022} on the few-shot MRQA benchmark (see below in section §\ref{ssec:baselines}) and found that MRQA performance is generally improved if RSS is used.
Therefore we use this model as basis for our MRQA models.
As template, we use \textit{context: <context> question: <question> answer: <mask>.}, and again use soft tokens (accounting for 9216 weights) which are optimized in addition to the full model during training.
The MRQA model is first trained on synthetic data for 1 epoch or at least 500 steps.
In a subsequent step, we further train on the annotated data from the few-shot splits.
For this, we use the hyperparameters reported in \citet{castel_how_2022}, that is a constant learning rate of 5e-5 for 512 training steps using the Adafactor optimizer \citep{shazeer_adafactor_2018} with a batch size of 32 and a dropout of 0.1.

We report the mean and standard deviation over 5 MRQA model runs while training the data generation model only once to save computation resources.

\begin{table*}[tb]
    \centering
    \begin{adjustbox}{max width=\textwidth}
        \begin{tabular}{l|c|c|c|c|c}
            \toprule
            \textbf{Model}                            & 0             & 16                    & 32                    & 64            & 128           \\ 
            \midrule
            Our approach                              & \textbf{85.5} & \textbf{86.4$\pm$0.6} & \textbf{88.3$\pm$0.4} & 87.7$\pm$0.4  & 89.3$\pm$0.6  \\
            Prompting+RSS Re-Impl                     & 71.5          & 84.0                  & 86.8                  & 86.8          & 88.8          \\
            Prompting+RSS~\citep{castel_how_2022}     & 71.4          & 85.6                  & 86.7                  & \textbf{87.9} & \textbf{89.4} \\
            Prompting~\citep{castel_how_2022}         & 60.0          & 82.6                  & 85.2                  & 86.7          & 89.0          \\
            Gotta~\citep{chen_gotta_2023}             & -             & 74.6$\pm$1.9          & 76.0$\pm$2.0          & 78.9$\pm$10.5 & 80.8$\pm$1.7  \\
            PMR (large)~\citep{xu_clozing_2022}       & 17.2          & 60.3$\pm$4.0          & 70.0$\pm$3.2          & 76.6$\pm$1.9  & 81.7$\pm$1.2  \\
            FewshotBARTL~\citep{chada_fewshotqa_2021} & -             & 68.9$\pm$2.7          & 72.3$\pm$1.0          & 73.6$\pm$1.9  & 79.4$\pm$1.5  \\
            Splinter (base)~\citep{ram_few-shot_2021} & -             & 54.6$\pm$6.4          & 59.2$\pm$2.1          & 65.2$\pm$1.4  & 72.7$\pm$1.0  \\ 
            Roberta (base)~\citep{ram_few-shot_2021}  & -             & 7.7$\pm$4.3           & 18.2$\pm$5.1          & 28.4$\pm$1.7  & 43.0$\pm$7.1  \\ 
            \bottomrule
        \end{tabular}
    \end{adjustbox}
    \caption{
        F1 score on SQuAD in the zero- and few-shot setting (16, 32, 64 \& 128 samples) for several existing models as well as our proposed data generation technique: Our approach yields a competitive result across all dataset sizes, but the settings with 16 and 32 labeled samples profit the most.
        Compared to our re-implementation of the Prompting+RSS model, our approach always performs better.
        Therefore, differences in implementation might result in slightly different numbers.
        Best model per dataset size marked \textbf{bold}.
    }
    \label{tab:results_squad}
\end{table*}

\subsection{Comparison Models}
\label{ssec:baselines}
We compare our approach to several recent and well performing MRQA models which we describe in the following.

\paragraph{Splinter}
This is the model proposed by \citet{ram_few-shot_2021} using an RSS pre-training phase which is fine-tuned on the few-shot MRQA datasets.
We show the results as reported by the authors for the base model.

\paragraph{FewshotBARTL}
FewshotBARTL is the best performing model reported in FewshotQA \citep{chada_fewshotqa_2021}.
This is a Prompting-based MRQA model using BART \citep{lewis_bart_2019}.

\paragraph{Prompting}
\citet{castel_how_2022} reported results of a Prompting-based MRQA model similar to FewshotBARTL but using T5 v1.1.

\paragraph{Prompting+RSS}
This model is similar to the Prompting model above, but with additional pre-training using RSS.
Since \citet{castel_how_2022} only report results on SQuAD, we consider this model only for RQ1.
For RQ2, we evaluate a re-implemented version of this model (see next model).

\paragraph{Prompting+RSS Re-Impl}
To account for implementational differences and to be able to evaluate the Prompting model with RSS on the full few-shot MRQA benchmark, we also consider a reimplementation of the Prompting+RSS model where we also directly perform MRQA via Prompting.
Additionally, we fine-tune soft tokens in the input initialized with weights from the embedding using the template.
To this end, we transform the sample into a prompt such that the pre-trained model answers the question using the given context.
This model is equal to the MRQA model we use in our approach, i.e. using the same template and the same hyperparameters.

\paragraph{Gotta}
This model proposed by \citet{chen_gotta_2023} is similar to Prompting with additional pre-training on entity-aware masks.

\paragraph{PMR}
PMR \citep{xu_clozing_2022} employs pre-training on automatically generated data in MRQA style and has a dedicated MRQA fine-tuning stage where the structure of inputs and outputs are similar.

\paragraph{Roberta}
We also show results reported by \citet{ram_few-shot_2021} for a model following the standard pre-training/fine-tuning paradigm using Roberta (base) \citep{liu_roberta_2019} with a span extraction head.

\section{Results and Discussion}

In order to judge the performance of our approach, we report the F1 score on the tested datasets for our approach as well as for the models we compare with.
We now examine our research questions using the reported results.

\begin{table}[tb]
    \centering
    \begin{adjustbox}{max width=\columnwidth}
        \begin{tabular}{l|c|c|c|c|c}
            \toprule
            \textbf{Model}            & 0    & 16   & 32   & 64   & 128  \\ 
            \midrule
            sampled answer            & 85.5 & 86.4 & 88.3 & 87.7 & 89.3 \\
            gold answers              & 87.3 & 87.6 & 89.5 & 88.3 & 90.3 \\
            \quad synthetic data only & 87.3 & 90.0 & 91.0 & 90.7 & 91.3 \\
            \bottomrule
        \end{tabular}
    \end{adjustbox}
    \caption{
        F1 score of an MRQA model using generated data from our approach on SQuAD in the zero- and few-shot setting (16, 32, 64 \& 128 samples) as well as comparing gold answers with sampled answers.
    }
    \label{tab:results_squad_zero_gold}
\end{table}

\subsection{RQ1: Synthetic Data Generation using LMs}
First, we answer the nether research questions in order to answer RQ1.
For this, we only evaluate on SQuAD.

\paragraph{RQ1.1: Synthetic Data for MRQA}
In general, as reported in table \ref{tab:results_squad}, our proposed method outperforms many existing approaches on SQuAD on all sizes although it does not perform best with 64 and 128 samples but is very close.
Also, there is a trend that more data improves our data generation approach although there is a fluctuation which we trace back to difficulties in training LMs for the MRQA task on little data in the final step as observed.

\paragraph{RQ1.2: Answer Selection}
For this research question, we compare the performance of generated data by our approach when using answers sampled using NER and when using the gold answers for SQuAD in table \ref{tab:results_squad_zero_gold}.
We can observe that generated data with sampled answers by NER performs in terms of F1 score on the MRQA model on average only 1.3\% worse than data generated using the gold answers.
Therefore the chosen answer sampling strategy can be a good replacement for answers as in the case of SQuAD.
Since we observed suboptimal training performance of the MRQA model on labeled data in the final step, we additionally report MRQA performance on synthetic data only (i.e., before fine-tuning on labeled data in a final step).
This shows that generated data can even perform better if care is taken when integrating the labeled samples into the eventual MRQA model.

\begin{table*}[tb]
    \centering
    \begin{adjustbox}{max width=\textwidth}
        \begin{tabular}{lHH|cH|cH|cH|cH|cH|cH|cH|c}
            \toprule
            \multirow{1}{*}{\textbf{Model}}           &                       &  & \multicolumn{2}{c|}{\textbf{TriviaQA}} & \multicolumn{2}{c|}{\textbf{NQ}} & \multicolumn{2}{c|}{\textbf{NewsQA}} & \multicolumn{2}{c|}{\textbf{SearchQA}} & \multicolumn{2}{c|}{\textbf{HotpotQA}} & \multicolumn{2}{c|}{\textbf{BioASQ}} & \multicolumn{2}{c|}{\textbf{TextbookQA}} & \textbf{Mean}                                                                                                 \\
            \midrule
            \multicolumn{9}{l}{\textit{16 Samples}}                                                                                                                                                                                                                                                                                                                                                                                                                                     \\
            \midrule
            Our approach                              & \textbf{86.4$\pm$0.6} &  & \textbf{76.4$\pm$0.5}                  &                                  & \textbf{68.5$\pm$0.6}                &                                        & \textbf{51.4$\pm$0.8}                  &                                      & 71.1$\pm$1.3                             &               & \textbf{72.4$\pm$0.6} &  & 72.3$\pm$1.5         &  & 60.1$\pm$2.4          &  & \textbf{67.5} \\
            Prompting+RSS Re-Impl                     & 84.0                  &  & 76.1                                   &                                  & 67.0                                 &                                        & 48.3                                   &                                      & \textbf{71.9}                            &               & 71.3                  &  & 73.0                 &  & \textbf{60.4}         &  & 66.9          \\
            Prompting~\citep{castel_how_2022}         & 82.6                  &  & 74.8                                   &                                  & 64.4                                 &                                        & 44.7                                   &                                      & 64.1                                     &               & 66.3                  &  & \textbf{74.7}        &  & 49.9                  &  & 62.7          \\
            Gotta~\citep{chen_gotta_2023}             & 74.6$\pm$1.9          &  & 63.3$\pm$8.0                           &                                  & 58.9$\pm$1.9                         &                                        & 47.3$\pm$2.5                           &                                      & 56.8$\pm$3.9                             &               & 59.8$\pm$2.1          &  & 66.1$\pm$3.1         &  & 38.5$\pm$5.3          &  & 55.8          \\
            PMR (large)~\citep{xu_clozing_2022}       & 60.3$\pm$4.0          &  & 56.2$\pm$3.1                           &                                  & 43.6$\pm$1.7                         &                                        & 30.1$\pm$3.7                           &                                      & 58.2$\pm$5.0                             &               & 46.1$\pm$4.7          &  & 54.2$\pm$3.4         &  & 31.0$\pm$1.8          &  & 45.6          \\
            FewshotBARTL~\citep{chada_fewshotqa_2021} & 68.9$\pm$2.7          &  & 65.2$\pm$1.8                           &                                  & 60.4$\pm$2.0                         &                                        & 48.4$\pm$2.2                           &                                      & 47.8$\pm$5.4                             &               & 58.0$\pm$1.8          &  & 63.0$\pm$1.1         &  & 37.7$\pm$3.7          &  & 54.4          \\
            Splinter (base)~\citep{ram_few-shot_2021} & 54.6$\pm$6.4          &  & 18.9$\pm$4.1                           &                                  & 27.4$\pm$4.6                         &                                        & 20.8$\pm$2.7                           &                                      & 26.3$\pm$3.9                             &               & 24.0$\pm$5.0          &  & 28.2$\pm$4.9         &  & 19.4$\pm$4.6          &  & 23.6          \\
            Roberta (base)~\citep{ram_few-shot_2021}  & 7.7$\pm$4.3           &  & 7.5$\pm$4.4                            &                                  & 17.3$\pm$3.3                         &                                        & 1.4$\pm$0.8                            &                                      & 6.9$\pm$2.7                              &               & 10.5$\pm$2.5          &  & 16.7$\pm$7.1         &  & 3.3$\pm$2.1           &  & 9.1           \\
            \midrule
            \multicolumn{9}{l}{\textit{32 Samples}}                                                                                                                                                                                                                                                                                                                                                                                                                                     \\
            \midrule
            Our approach                              & \textbf{88.3$\pm$0.4} &  & \textbf{76.8$\pm$0.5}                  &                                  & \textbf{68.5$\pm$0.8}                &                                        & 50.6$\pm$0.8                           &                                      & \textbf{72.9$\pm$0.8}                    &               & \textbf{73.4$\pm$0.6} &  & 74.5$\pm$0.8         &  & \textbf{61.0$\pm$1.8} &  & \textbf{68.2} \\
            Prompting+RSS Re-Impl                     & 86.8                  &  & 75.6                                   &                                  & 64.0                                 &                                        & 49.0                                   &                                      & 71.1                                     &               & 71.7                  &  & 73.1                 &  & \textbf{61.0}         &  & 66.5          \\
            Prompting~\citep{castel_how_2022}         & 85.2                  &  & 74.8                                   &                                  & 66.7                                 &                                        & 48.8                                   &                                      & 66.2                                     &               & 70.3                  &  & \textbf{76.8}        &  & 51.2                  &  & 65.0          \\
            Gotta~\citep{chen_gotta_2023}             & 76.0$\pm$2.0          &  & 61.9$\pm$4.8                           &                                  & 59.8$\pm$2.4                         &                                        & \textbf{51.2$\pm$1.5}                  &                                      & 63.1$\pm$3.1                             &               & 62.7$\pm$1.2          &  & 69.5$\pm$1.0         &  & 46.3$\pm$3.7          &  & 59.2          \\
            PMR (large)~\citep{xu_clozing_2022}       & 70.0$\pm$3.2          &  & 66.3$\pm$2.5                           &                                  & 48.5$\pm$3.5                         &                                        & 36.6$\pm$2.1                           &                                      & 64.8$\pm$2.2                             &               & 52.9$\pm$2.5          &  & 62.9$\pm$2.4         &  & 36.4$\pm$3.2          &  & 52.6          \\
            FewshotBARTL~\citep{chada_fewshotqa_2021} & 72.3$\pm$1.0          &  & 65.1$\pm$1.2                           &                                  & 61.5$\pm$1.7                         &                                        & 51.7$\pm$1.7                           &                                      & 58.3$\pm$1.5                             &               & 60.4$\pm$0.2          &  & 67.8$\pm$1.0         &  & 37.7$\pm$9.8          &  & 57.5          \\
            Splinter (base)~\citep{ram_few-shot_2021} & 59.2$\pm$2.1          &  & 28.9$\pm$3.1                           &                                  & 33.6$\pm$2.4                         &                                        & 27.5$\pm$3.2                           &                                      & 34.8$\pm$1.8                             &               & 34.7$\pm$3.9          &  & 36.5$\pm$3.2         &  & 27.6$\pm$4.3          &  & 31.9          \\
            Roberta (base)~\citep{ram_few-shot_2021}  & 18.2$\pm$5.1          &  & 10.5$\pm$1.8                           &                                  & 22.9$\pm$0.7                         &                                        & 3.2$\pm$1.7                            &                                      & 13.5$\pm$1.8                             &               & 10.4$\pm$1.9          &  & 23.3$\pm$6.6         &  & 4.3$\pm$0.9           &  & 12.6          \\
            \midrule
            \multicolumn{9}{l}{\textit{64 Samples}}                                                                                                                                                                                                                                                                                                                                                                                                                                     \\
            \midrule
            Our approach                              & 87.7$\pm$0.4          &  & 76.2$\pm$0.5                           &                                  & \textbf{70.4$\pm$0.4}                &                                        & \textbf{56.4$\pm$0.7}                  &                                      & \textbf{75.0$\pm$1.4}                    &               & \textbf{74.7$\pm$0.2} &  & 76.8$\pm$0.5         &  & \textbf{64.0$\pm$1.0} &  & \textbf{70.5} \\
            Prompting+RSS Re-Impl                     & 86.8                  &  & \textbf{76.3}                          &                                  & 68.7                                 &                                        & 52.9                                   &                                      & 72.5                                     &               & 73.4                  &  & 78.5                 &  & 61.8                  &  & 69.2          \\
            Prompting~\citep{castel_how_2022}         & 86.7                  &  & 75.3                                   &                                  & 68.5                                 &                                        & 49.9                                   &                                      & 71.7                                     &               & 73.1                  &  & \textbf{80.4}        &  & 55.6                  &  & 67.8          \\
            Gotta~\citep{chen_gotta_2023}             & 78.9$\pm$10.5         &  & 59.6$\pm$11.9                          &                                  & 63.6$\pm$11.0                        &                                        & 54.3$\pm$13.0                          &                                      & 66.3$\pm$12.5                            &               & 64.3$\pm$11.7         &  & 73.2$\pm$11.5        &  & 51.2$\pm$12.8         &  & 61.8          \\
            PMR (large)~\citep{xu_clozing_2022}       & 76.6$\pm$1.9          &  & 67.5$\pm$1.7                           &                                  & 53.4$\pm$2.3                         &                                        & 46.8$\pm$2.6                           &                                      & 69.3$\pm$2.4                             &               & 61.7$\pm$2.1          &  & 71.5$\pm$1.8         &  & 43.4$\pm$3.6          &  & 59.1          \\
            FewshotBARTL~\citep{chada_fewshotqa_2021} & 73.6$\pm$1.9          &  & 64.6$\pm$1.4                           &                                  & 63.0$\pm$2.1                         &                                        & 53.5$\pm$0.9                           &                                      & 65.5$\pm$2.4                             &               & 62.9$\pm$1.6          &  & 73.9$\pm$0.8         &  & 45.0$\pm$1.7          &  & 61.2          \\
            Splinter (base)~\citep{ram_few-shot_2021} & 65.2$\pm$1.4          &  & 35.5$\pm$3.7                           &                                  & 38.2$\pm$2.3                         &                                        & 37.4$\pm$1.2                           &                                      & 39.8$\pm$3.6                             &               & 45.4$\pm$2.3          &  & 49.5$\pm$3.6         &  & 35.9$\pm$3.1          &  & 40.2          \\
            Roberta (base)~\citep{ram_few-shot_2021}  & 28.4$\pm$1.7          &  & 12.5$\pm$1.4                           &                                  & 24.2$\pm$1.0                         &                                        & 4.6$\pm$2.8                            &                                      & 19.8$\pm$2.4                             &               & 15.0$\pm$3.9          &  & 34.0$\pm$1.8         &  & 5.4$\pm$1.1           &  & 16.5          \\
            \midrule
            \multicolumn{9}{l}{\textit{128 Samples}}                                                                                                                                                                                                                                                                                                                                                                                                                                    \\
            \midrule
            Our approach                              & 89.3$\pm$0.6          &  & \textbf{77.4$\pm$0.3}                  &                                  & \textbf{73.0$\pm$0.4}                &                                        & \textbf{57.4$\pm$0.7}                  &                                      & \textbf{78.6$\pm$0.5}                    &               & \textbf{76.5$\pm$0.2} &  & 82.4$\pm$0.3         &  & \textbf{63.5$\pm$1.6} &  & \textbf{72.7} \\
            Prompting+RSS Re-Impl                     & 88.8                  &  & 77.1                                   &                                  & 71.3                                 &                                        & 55.1                                   &                                      & 75.9                                     &               & 75.9                  &  & 81.8                 &  & 62.4                  &  & 71.4          \\
            Prompting~\citep{castel_how_2022}         & 89.0                  &  & 76.7                                   &                                  & 69.9                                 &                                        & 51.8                                   &                                      & 73.4                                     &               & 74.6                  &  & \textbf{85.2}        &  & 58.0                  &  & 69.9          \\
            Gotta~\citep{chen_gotta_2023}             & 80.8$\pm$1.7          &  & 60.0$\pm$3.6                           &                                  & 64.9$\pm$1.2                         &                                        & \textbf{57.4$\pm$1.2}                  &                                      & 69.8$\pm$1.5                             &               & 66.7$\pm$1.8          &  & 78.6$\pm$2.1         &  & 53.3$\pm$1.7          &  & 64.4          \\
            PMR (large)~\citep{xu_clozing_2022}       & 81.7$\pm$1.2          &  & 70.3$\pm$0.5                           &                                  & 57.4$\pm$2.6                         &                                        & 52.3$\pm$1.4                           &                                      & 70.0$\pm$1.1                             &               & 65.9$\pm$1.0          &  & 78.8$\pm$0.5         &  & 45.1$\pm$1.2          &  & 62.8          \\
            FewshotBARTL~\citep{chada_fewshotqa_2021} & 79.4$\pm$1.5          &  & 65.8$\pm$0.9                           &                                  & 64.3$\pm$1.3                         &                                        & 57.0$\pm$0.9                           &                                      & 67.7$\pm$1.0                             &               & 75.1$\pm$1.5          &  & 75.0$\pm$1.5         &  & 48.4$\pm$2.7          &  & 64.8          \\
            Splinter (base)~\citep{ram_few-shot_2021} & 72.7$\pm$1.0          &  & 44.7$\pm$3.9                           &                                  & 46.3$\pm$0.8                         &                                        & 43.5$\pm$1.3                           &                                      & 47.2$\pm$3.5                             &               & 54.7$\pm$1.4          &  & 63.2$\pm$4.1         &  & 42.6$\pm$2.5          &  & 48.9          \\
            Roberta (base)~\citep{ram_few-shot_2021}  & 43.0$\pm$7.1          &  & 19.1$\pm$2.9                           &                                  & 30.1$\pm$1.9                         &                                        & 16.7$\pm$3.8                           &                                      & 27.8$\pm$2.5                             &               & 27.3$\pm$3.9          &  & 46.1$\pm$1.4         &  & 8.2$\pm$1.1           &  & 25.0          \\
            \midrule
            \multicolumn{9}{l}{\textit{Full Dataset}}                                                                                                                                                                                                                                                                                                                                                                                                                                   \\
            \midrule
            Splinter (base)~\citep{ram_few-shot_2021} & 92.2                  &  & 76.5                                   &                                  & 81.0                                 &                                        & 71.3                                   &                                      & 83.0                                     &               & 80.7                  &  & 91.0\footnotemark[1] &  & 54.5\footnotemark[1]  &  & 76.9          \\
            Roberta (base)~\citep{ram_few-shot_2021}  & 90.3                  &  & 74.0                                   &                                  & 79.6                                 &                                        & 69.8                                   &                                      & 81.5                                     &               & 78.7                  &  & 84.1\footnotemark[1] &  & 35.8\footnotemark[1]  &  & 71.9          \\
            \bottomrule
        \end{tabular}
    \end{adjustbox}
    \caption{
        Results of several models as evaluated by means of the F1 score for the few-shot MRQA benchmark excluding SQuAD.
        Our approach performs best across various dataset domains and sizes, and even outperforms the full data setting for TriviaQA and TextbookQA.
        Best model per dataset domain and size marked \textbf{bold}.\hfill$^1$ 1024 training samples
    }
    \label{tab:results_all}
\end{table*}

\paragraph{RQ1.3: Zero-shot Performance}
As reported in table \ref{tab:results_squad}, without any labeled data our approach gets an F1 score of 85.5\%.
Although the performance increases if labeled data is added, this clearly outlines that there are strong zero-shot capabilities by employing data generation for MRQA.
Therefore, we can conclude that the LM has learned during pre-training the relationship between questions and answers to an extent that is useful for SQuAD.

In summary, as an answer to RQ1, our Prompting-based data generation approach efficiently employs LMs for MRQA increasing performance compared to existing work.
The proposed method proves to be competitive on SQuAD and performs well especially without training data, establishing a new state of the art.

\subsection{RQ2: Domain Generalization}

\begin{figure}[h]
    \centering
    \includegraphics[trim=20 25 50 40,clip,width=\linewidth,keepaspectratio]{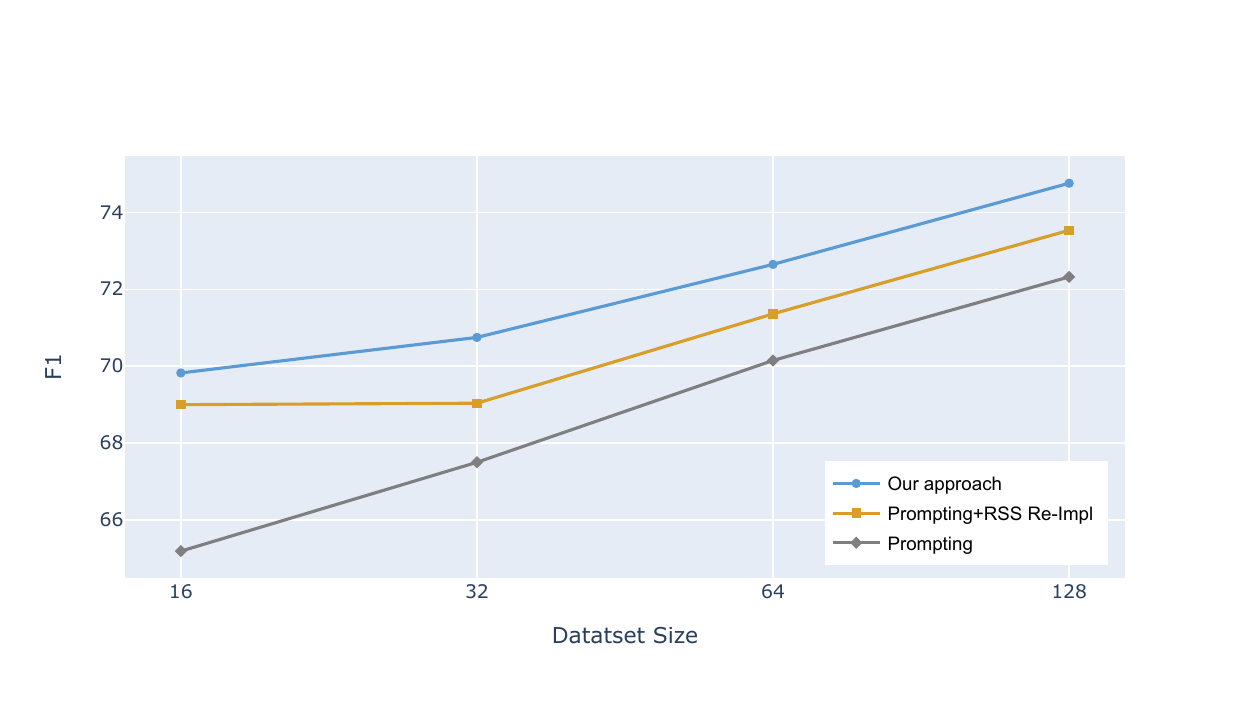}
    \caption{
        MRQA performance (F1) as a function of dataset sizes for the best performing approaches on the mean of all datasets in the few-shot MRQA benchmark.
    }
    \label{fig:plot_mean}
\end{figure}

To answer the second research question, we report results on the few-shot MRQA benchmark excluding SQuAD in table \ref{tab:results_all} (we additionally show the results of the best performing approaches for the mean of all datasets including SQuAD and all dataset sizes in figure \ref{fig:plot_mean}).
For NQ, HotpotQA and TextbookQA we consistently rank first across all dataset sizes.
Largest absolute increases of F1 score can be observed on NQ with 32 samples (1.8\%), SearchQA with 32, 64 and 128 samples (1.8\%, 2.5\% and 2.7\%, respectively), HotpotQA with 32 samples (1.7\%), and TextbookQA with 64 samples (2.2\%).
Also, we see that in general the performance increases with more labeled data, although this behavior is not consistent in the case of TriviaQA, NewsQA and TextbookQA.

Interestingly, on all sizes of BioASQ, our approach performs worse than directly using a Prompting model.
Since the same applies to the Prompting approach using RSS, we assume that the MRQA model in our approach also suffers from the RSS pre-training although we cannot find reasons for RSS performing worse in this domain.

Finally, we note that we saw quite high fluctuations between model training runs in the few-shot setting.
We assume this is owed to suboptimal hyperparameters which do not generalize well across domains and to too few samples.
Training on 128 samples or less can easily lead to overfitting resulting in loss of generalization.
Therefore, with a better incorporation of few labeled data into the models, we believe that the MRQA performance of a Prompting-based data generation approach can further be increased.

For answering research question RQ2, we can conclude that our data generation approach also generalizes to other domains as demonstrated by the few-shot MRQA benchmark.
To analyze the benefit of our approach, we further investigate the quality of the generated question-answer pairs.



\section{Analysis}
\begin{figure}[t]
    \centering
    \includegraphics[width=\linewidth,keepaspectratio]{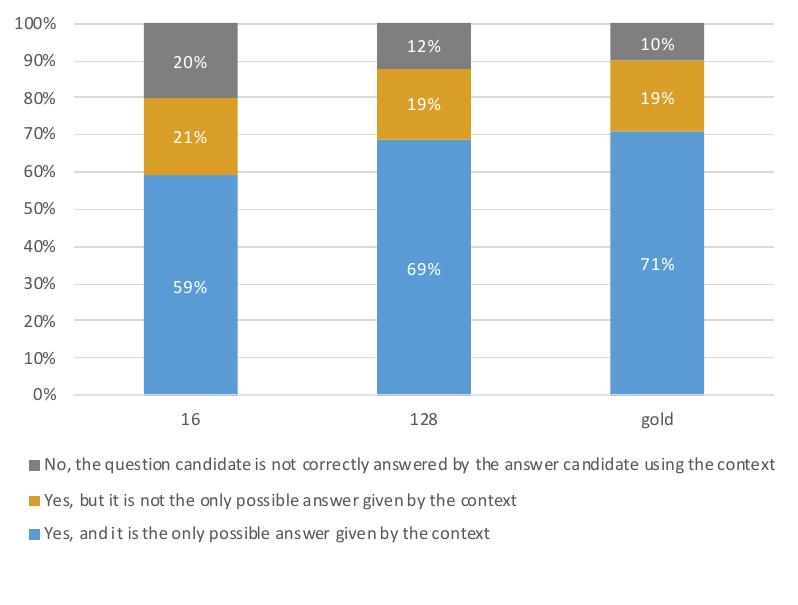}
    \caption{
        For the NewsQA dataset, 100 question-answer pairs were quality-assessed by humans (question: "Is the question candidate correctly answered by the answer candidate?") in each setting (generated data taking 16 and 128 samples into account as well as labeled (gold) data).
    }
    \label{fig:question_analysis}
\end{figure}

In order to assess the data quality of the generated questions and answers a user study was taken.
A total of 30 people, which were recruited via the Prolific platform, took part in the study.
In order to achieve a high significance, the selection of participants was restricted by the following screening:
The participants must have a bachelor's degree or higher, speak English as their primary language, and have a 100\% approval rate with Prolific.
The aim of the study is to find out whether the generated data from our approach provide a comparable data quality as labeled data, are correct question-answer pairs with respect to the context, and can be improved quality-wise with more labeled samples.

In order to carry out the analysis, the study participants were randomly given 10 samples per participant.
In these, the context, the question candidate as well as the answer candidate was provided.
In total 300 question-answer pairs were individually assessed by humans with regard to their quality.
For this purpose, one question was asked per shown sample about the data quality\footnote{The participants were asked to answer the following question: "Is the question candidate correctly answered by the answer candidate?"}, which made it possible to distinguish between correct answers, partially correct answers (in the case of several answers possible from the context) as well as incorrect answers.
In the introduction it was explicitly pointed out that only the given context may be considered for answering the question candidate.

The dataset used was NewsQA, which has the lowest F1 compared to the other datasets used (cf. table \ref{tab:results_all}) and the quality of the generated question-answer pairs is thus tested on a comparatively difficult domain.
To achieve the above study objectives, 3 different settings were chosen, each given to 10 participants: generated data using 16 and 128 samples as well as gold data.

The results of the study are shown in figure \ref{fig:question_analysis}.
First, we can observe that with 128 labeled samples, generated data is comparable to the quality of labeled data.
Although with only 16 samples the majority of generated question-answer pairs is of high quality (59\%) despite the extremely small effort needed by humans, this lacks behind 10 points in absolute percentage when compared to the data quality generated using 128 samples.
Therefore, labeling 128 samples can be sufficient for our approach for the NewsQA dataset to get a similar quality of question-answer pairs compared to human annotated data that is more complex and costly.

\section{Conclusion}
In summary, we introduced a new approach for MRQA that makes use of the linguistic knowledge encoded in LMs.
To this end, we proposed to generate synthetic question-answer pairs for MRQA and run several experiments to test the performance of our approach in the zero- and few-shot setting, thereby also showing its generalizability.
As a result, we have shown that LMs can be more effectively used, and find that our approach outperforms many state-of-the-art approaches for the MRQA task.
Furthermore, in some settings, synthetic data is even on par with human annotated data.
However, the performance heavily depends on the domain under consideration, with the highest absolute increase of performance for the most difficult domain.
Finally, we demonstrated in a user study that it is possible with only taking into account 128 human annotated samples to generate question-answer pairs which are comparable to human annotated data in terms of quality.
We believe that there are many more ways to effectively use LMs and hope that our work will be an incentive to explore other possibilities.

\section{Future Work}
Although we have shown that our approach using NER sampled answers performs comparably well, other methods are worth to be explored too.
For example, leveraging LLMs to also generate the answer is interesting, but poses additional challenges for extractive MRQA.
They struggle in providing the start and end indices of answers if they are also used for selecting the answers, which renders some model architectures invalid.
We therefore believe that more investigation is necessary to enable more effective and more efficient use of LLMs.
For the generation, feedback (for example provided by humans) could be included to continuously improve the quality of synthetic data.

Regarding the MRQA model, other methods to incorporate synthetic data should be taken into account too.
For example, adopting in-context learning for extractive MRQA is a highly interesting direction but out of scope of this work.

Furthermore, since our approach performs well for SQuAD in the zero-shot setting as well, it should further be investigated, also for other domains.


\section{Ethical Considerations}
Regarding our user study, participants were acquired using the Prolific platform.
Prior to obtaining consent, we provided detailed instructions and descriptions of how their answers are processed, and that their participation is voluntary.
We also ensured that payment was not below Prolific's recommendation for the participants.
Regarding privacy, we did not collect any personal or identifying data, or data other than for the question mentioned in this work.

\section*{Acknowledgements}

This work is supported by IBM Research AI through the IBM AI Horizons Network.

%
\section{References}\label{reference}

\bibliographystyle{lrec-coling2024-natbib}
\bibliography{custom}


\end{document}